\documentclass[10pt]{article}

\PassOptionsToPackage{numbers}{natbib}

\usepackage[utf8]{inputenc}

\usepackage{fullpage}

\usepackage{amsmath}
\usepackage{amssymb}

\usepackage{amssymb,euscript,latexsym,amsmath}
\usepackage{xcolor}
\usepackage{graphicx} 
\usepackage{epstopdf}
\usepackage{asymptote}
\usepackage{bm} 
\usepackage{enumerate} 
\usepackage{color} 
\usepackage{setspace} 
\usepackage{subcaption} 
\usepackage{booktabs}
\usepackage[
pdftitle={NQS with RetNets},
pdfcreator={pdflatex},
pdfsubject={preprint},
hyperindex = {true},
colorlinks = {true},
linkcolor = {blue},
citecolor = {blue}
]{hyperref}
\usepackage{array}
\usepackage{cancel}
\usepackage{stmaryrd}

\usepackage[dvips]{epsfig}
\usepackage{epstopdf}
\usepackage{setspace}
\usepackage{lipsum}
\usepackage[shortlabels]{enumitem}
\usepackage{algorithm, algpseudocode}

\usepackage{cleveref}
\crefformat{equation}{(#2#1#3)}

\usepackage[preprint]{neurips_2024}

\newcommand{\abs}[1]{\left\lvert #1 \right\rvert}

\newcommand{\ip}[1]{\left\langle #1 \right\rangle}
\newcommand{\paren}[1]{\left( #1 \right)}
\newcommand{\braces}[1]{\left\{ #1 \right\}}
\newcommand{\bracket}[1]{\left[ #1 \right]}
\newcommand{\bra}[1]{\left\langle #1 \right\rvert}
\newcommand{\ket}[1]{\left\lvert #1 \right\rangle}

\newcommand\numberthis{\addtocounter{equation}{1}\tag{\theequation}}

\title{Retentive Neural Quantum States: Efficient Ans\"atze\\ for
Ab Initio Quantum Chemistry}

\author{%
  Oliver Knitter$^{1,2, *}$\quad
  Dan Zhao$^{2,3}$\quad
  James Stokes$^{1}$\quad
  Martin Ganahl$^{2}$\\
  \textbf{Stefan Leichenauer$^{2}$\quad
  Shravan Veerapaneni$^{1}$}\\
  $^1$University of Michigan\quad
  $^2$SandboxAQ\quad
  $^3$New York University\\
  \texttt{knitter@umich.edu}
}

\begin{document}
\maketitle

\vspace{-2em}

\begin{abstract}
   Neural-network quantum states (NQS) has emerged as a powerful application of quantum-inspired deep learning for variational Monte Carlo methods, offering a competitive alternative to existing techniques for identifying ground states of quantum problems. A significant advancement toward improving the practical scalability of NQS has been the incorporation of autoregressive models, most recently transformers, as variational ansatze. Transformers learn sequence information with greater expressiveness than recurrent models, but at the cost of increased time complexity with respect to sequence length.
    We explore the use of the retentive network (RetNet), a recurrent alternative to transformers, as an ansatz for solving electronic ground state problems in \textit{ab initio} quantum chemistry. Unlike transformers, RetNets overcome this time complexity bottleneck by processing data in parallel during training, and recurrently during inference. We give a simple computational cost estimate of the RetNet and directly compare it with similar estimates for transformers, establishing a clear threshold ratio of problem-to-model size past which the RetNet's time complexity outperforms that of the transformer. Though this efficiency can comes at the expense of decreased expressiveness relative to the transformer, we overcome this gap through training strategies that leverage the autoregressive structure of the model---namely, variational neural annealing. Our findings support the RetNet as a means of improving the time complexity of NQS without sacrificing accuracy. We provide further evidence that the ablative improvements of neural annealing extend beyond the RetNet architecture, suggesting it would serve as an effective general training strategy for autoregressive NQS.
\end{abstract}


\section{Introduction}
\label{sec:intro}

Variational Monte Carlo (VMC) with neural network quantum states (NQS) is a generative deep learning framework \cite{carleo2017solving} that utilizes a classical neural network to efficiently encode the state vector of a quantum many-body system. This neural network ansatz learns a state vector approximating the ground state eigenvector of a given quantum Hamiltonian $H$, via a VMC training algorithm analogous to policy gradient methods in reinforcement learning. More specifically, for state vectors formulated in the Hilbert space $\mathbb{C}^{2^n}$, binary strings of length $n$ are sampled from the Born probability distribution associated with the approximate state vector. These samples are used to efficiently estimate both the expectation value of $H$ and the associated network parameter gradients. Thus NQS bypasses the need to explicitly store the full ansatz state vector or Hamiltonian matrix, whose dimensions grow exponentially with respect to $n$. So long as $H$ is sufficiently row sparse with easily retrievable entries---a loose restriction for Hamiltonians relevant to many practical applications---the entire algorithm will execute in polynomial time with respect to $n$.

Direct conceptual relationships have been drawn between NQS, as a high-level numerical black box solver, and both natural evolution strategies \cite{zhao2020natural} and variational quantum eigensolvers (VQEs). In latter case, NQS may be viewed not only as an analogous method but also as a direct pipeline for de-quantization of existing VQEs \cite{knitter2022vnls}, which rely on parameterized quantum circuits to prepare physical trial wavefunctions. As NQS is fundamentally problem-agnostic, a property it shares with VQEs, it is unsurprising that NQS has been successfully demonstrated on a variety of problems: Ising models \cite{carleo2017solving}, combinatorial optimization \cite{gomes2019classical, zhao2020natural}, and high-dimensional linear algebra \cite{knitter2022vnls}. While some of the existing demonstrations of these use cases are proofs of concept, they all depict a significant greater potential for NQS to perform accurately and efficiently on practically-motivated problems.

An exciting area of application of significant recent interest is to use NQS for \textit{ab initio} quantum chemistry, specifically to solve electronic structure problems \cite{Choo_2020, barrett2022autoregressive, zhao2023nnqsmade, li2024improved, wu2023nnqstransformer}. In this application, the second quantized electronic structure Hamiltonian corresponding to some molecule, which encodes the energy of the joint wavefunction describing all of that molecule's electrons, is mapped to a qubit-based Hamiltonian that can be targeted by NQS. This line of research parallels similar recent efforts to address these problems using VQEs \cite{cerezo2021variational, nam2020ground, tilly2022variational}, and has positioned NQS as a potentially viable alternative to more established methods like coupled cluster \cite{bartlett2007coupled} and tensor network methods like DMRG \cite{schollwock2005density}. 

The first applications of NQS to electronic structure calculations utilized restricted Boltzmann machines---which model unnormalized wave functions that necessitate approximate Monte Carlo sampling methods---as ansatz networks \cite{Choo_2020}. Although there is promising research being conducted with unnormalized wave functions \cite{li2024improved}, much of NQS research in this area now relies on autoregressive wavefunction ansatze \cite{barrett2022autoregressive}, which are normalized by construction and furthermore allow for exact sampling from their Born probability distribution. This paradigm has resulted in scalable quantum chemistry solvers, performing with accuracies comparable to CCSD, using the masked feedforward network MADE \cite{zhao2023nnqsmade}, and---more recently---transformers \cite{wu2023nnqstransformer}. Transformers constitute a compelling alternative to MADE since they are more parameter-efficient and do not explicitly grow in size with respect to the number of qubits $n$, but they do possess one inherent computational drawback: the cost of performing a forward pass grows quadratically with the qubit number both during training and inference. When employed in large language model (LLM) applications, it is understood that this quadratic scaling does not contribute significantly to the compute requirements for training transformer-based models, as reflected in standard floating point operation (FLOP) count estimates \cite{kaplan2020scaling}. For NQS, however, each optimization step necessitates multiple forward passes of the ansatz in order to generate Monte Carlo samples, so quadratic cost poses a nonneglible computational bottleneck at scale.

In this work, we introduce a novel NQS ansatz based on retentive networks (RetNets) \cite{sun2023retnet}, a recurrent model designed as an alternative to transformers for LLM applications. Unlike transformers, RetNets process inputs in parallel during training, and recurrently during inference, allowing for a linear inference cost with respect to $n$. Since RetNets can alleviate the inference cost of transformers and have been shown to exhibit baseline performance comparable to transformers for natural language processing tasks, they serve as a promising candidate for NQS-based electronic structure calculations. 

\paragraph{Contributions.} This paper aims to expose this potential as follows: \cref{sec:background_nqs} provides the relevant background information and context for this work: the application of autoregressive NQS to \textit{ab initio} quantum chemistry, the principles of variational neural annealing, the NQS-Transformer ansatz, and the RetNet architecture. We also extend existing transformer FLOP count estimates \cite{kaplan2020scaling} to RetNets, demonstrating how the latter can provide substantial FLOP savings under reasonable model constraints. In \cref{sec:architecture}, we describe the primary contributions of this work:
\begin{itemize}
    \item We present the RetNet as a viable drop-in replacement to the transformer in NQS, illustrating how the structure and demands of NQS make it amenable to such a replacement.

    \item We demonstrate how variational neural annealing \cite{hibat2021variational} increases the robustness of NQS performance with respect to choice of training hyperparameters.
\end{itemize}
Section \ref{sec:experiments} details experimental results that demonstrate the efficacy of RetNets as NQS ansatze on a slate of fundamental baseline molecules. We also experimentally demonstrate the beneficial effects of variational neural annealing on the training of NQS ansatze, by showing how annealing can improve the accuracy of NQS for these molecules, using model sizes that are smaller than what is previously reported in the literature. When shown alongside more incremental architectural changes intended to reduce the computational cost of NQS, these results indicate several promising avenues for improving the practical operation of NQS at scale. We give a conclusive summary of this work, alongside some discussion of potential future directions for it, in \cref{sec:conclusions}.

\section{Related Work}
The first applications of NQS to \textit{ab initio} quantum chemistry utilized unnormalized RBMs \cite{Choo_2020}, whose state distributions cannot be sampled exactly. In parallel with the development of autoregressive NQS, a separate line of research considers a non-stochastic optimization procedure to make RBMs viable as quantum chemistry ansatze \cite{li2024improved}. This optimization deterministically selects qubit spin configurations according to a relevance score based on how different configurations are related to each other under the problem Hamiltonian, as formulated below in \cref{eqn:pauli_ham}. Bypassing the need to sample from the RBM's unnormalized state distribution, non-stochastic NQS has been successfully demonstrated at the same scales as those targeted by autoregressive NQS. Both these forms of NQS exhibit strong potential and warrant further exploration on their own merits---a deeper understanding of the benefits and limitations of each method is likely necessary to make a decisive comparison between them.

Another line of research in autoregressive NQS explores the incorporation of quantum number symmetries within the ansatz \cite{malyshev2023autoregressive}.
When properly encoded into the NQS framework, physical symmetries are helpful for constraining the NQS ansatz to physically plausible subspaces of the overall state space.

Considering the main example of this work, the second quantized Hamiltonian of an individual molecule is defined by the geometric arrangement and electric charges of the molecule's nuclei, and by the number and form of the spin-orbitals comprising its basis set. The number of electrons in the molecule, a fixed parameter of the problem, does not contribute to the Hamiltonian's construction. Without further constraint, NQS will search over the space generated by all possible spin-orbital occupancies, regardless of electron count. For ease of training, autoregressive NQS typically enforces a hard constraint on the ansatz to ensure it only searches the subspace generated by occupancies with the correct number of electrons \cite{zhao2023nnqsmade}. The successful incorporation of additional physical symmetries may further reduce the number of iterations required for effective training \cite{malyshev2023autoregressive}.

\section{Preliminaries}
\label{sec:background_nqs}

\subsection{Neural Quantum States for \textit{ab initio} Quantum Chemistry}
NQS uses a classical neural network to represent the (normalized) quantum state vector $\ket{\psi_\theta}\in\mathbb{C}^{2^n}$ of an $n$-qubit wavefunction. This neural network ansatz accepts standard basis indices (encoded in binary as qubit spins) as inputs, and returns the logarithm of the corresponding state vector component,
\begin{equation}
    \log\ip{x|\psi_\theta} = \log\psi_\theta(x):\braces{0,1}^n\rightarrow\mathbb{C},\;\;\;\theta\in\mathbb{R}^d
    \label{eqn:nqs_ansatz}
\end{equation}
This ansatz is used to evaluate the quantum expectation value of a Hamiltonian, which is a Hermitian operator $H$ acting on the $2^n$-dimensional space containing $\ket{\psi_\theta}$. Like the VQE, NQS seeks to minimize the quantum expectation value
\begin{equation}
    L(\theta) = \bra{\psi_\theta}H\ket{\psi_\theta}
    \label{eqn:cost_fn}
\end{equation}
in order to estimate the ground state of $H$.

In the context of this paper, $H$ is taken to be second quantized electronic Hamiltonian of an isolated molecule, comprised of the kinetic and potential energies of the joint electronic wavefunction, treating the atomic nuclei as singular classical point charges, an assumption known as the Born-Oppenheimer approximation \cite{born1927quantum}
\begin{equation}
    H_{\text{e}} = -\frac{1}{2}\sum_j\Delta_j - \sum_{j,J}\frac{Z_J}{\abs{r_j - R_J}} + \frac 1 2 \sum_{j\neq k} \frac 1 {\abs{r_j-r_k}}
    \label{eqn:atomic_ham}
\end{equation}
Here $Z_J$ and $R_J$ represent the atomic number and position for atomic nucleus $J$, and $r_j$ represents the position of electron $j$.

Fixing a basis set of $n$ spin-orbitals, the second quantized Hamiltonian $H$ acts on a $2^n$-dimensional state space defined by possible spin-orbital occupancies. Using the Jordan-Wigner transformation, $H$ is further mapped to a Hamiltonian on an $n$-qubit state space, amenable to NQS. More specifically, the Jordan-Wigner transformation expresses $H$ as a real linear combination of Pauli strings:
\begin{equation}
\label{eqn:pauli_ham}
    H = \sum_j \alpha_j \paren{\sigma^j_1\otimes\dots\otimes\sigma^j_n},\;\;\;\sigma^j_k\in\braces{I,X,Y,Z}
\end{equation}

The operators $\braces{I,X,Y,Z}$ form an orthogonal set under the Hilbert-Schmidt inner product, which spans the real vector space of $2\times2$ Hermitian matrices. Taking all tensor products of $n$ copies of them produces an analogous basis for the space of $2^n\times2^n$ Hermitian matrices: the space of possible $n$-qubit Hamiltonians. The scalars $\alpha_i$ in \cref{eqn:pauli_ham} derive from the Hilbert-Schmidt decomposition of $H$ in this basis. Although the $n$-qubit Hamiltonian space is $4^n$-dimensional, second quantization and Jordan-Wigner together guarantee that the number of terms in \cref{eqn:pauli_ham} is $O(n^4)$.

These local Pauli strings correspond with fundamental quantum gates, and by applying these gates on a physically prepared trial wavefunction $\ket{\psi_\theta}$, and performing quantum measurements on the result, VQEs can generate stochastic estimates of \cref{eqn:cost_fn} \cite{tilly2022variational}. In contrast, NQS constructs a different set of stochastic estimates, based on the following estimator,
\begin{align*}
    L(\theta) &= \sum_x \abs{\ip{x|\psi_\theta}}^2 \frac{\bra{x}H\ket{\psi_\theta}}{\ip{x|\psi_\theta}}\\
    &=\underset{x\sim\abs{\psi_\theta}^2}{\mathbb{E}}\bracket{l_\theta(x)},\;\;\;l_\theta(x) = \frac{\bra{x}H\ket{\psi_\theta}}{\ip{x|\psi_\theta}}
    \numberthis
    \label{eqn:nqs_loss}
\end{align*}

This formulation recasts \cref{eqn:cost_fn} into the expected value of $l_\theta(x)$, called the \textit{local energy}, where $x\sim\abs{\psi_\theta(x)}^2$. Thus we may approximate \cref{eqn:cost_fn} by sampling basis vectors from the probability distribution $\abs{\psi_\theta}^2$, and averaging their local energies. Similarly, gradients can be estimated via a variant on the log-derivative trick \cite{mohamed2020monte},
\begin{equation}
    \label{eqn:nqs_gradient}
    \nabla_\theta L(\theta) =  2\operatorname{Re}\underset{x\sim\abs{\psi_\theta}^2}{\mathbb{E}}\bracket{\paren{l_\theta(x) - b}\nabla_\theta\log\ip{\psi_\theta|x}}
\end{equation} 
The scalar $b$ in \cref{eqn:nqs_gradient} is a hyperparameter traditionally chosen to equal $L(\theta)$ \cite{barrett2022autoregressive, carleo2017solving} to reduce variance, but changing $b$ introduces no bias to the gradient estimator \cite{mohamed2020monte}.

By relying on the stochastic formulations from \cref{eqn:nqs_loss} and \cref{eqn:nqs_gradient}, NQS remains computationally feasible if $\ket{\psi_\theta}$ and $H$ satisfy two key conditions discussed in \cite{nest2009simulating}: 
\begin{enumerate}
    \item $H$ is an efficiently computable operator: each row of $H$ must have, with respect to $n$, polynomially many entries, with the location and value of each entry computable in polynomial time \cite{knitter2022vnls}.
    \item $\ket{\psi_\theta}$ models classically tractable (CT) states: its individual entries $\ip{x|\psi_\theta}$ must be efficiently retrievable, and 
    it must also be possible to efficiently sample spin configurations from $x\sim\abs{\psi_\theta}^2$.
\end{enumerate}
Let us consider condition 1. Since Hamiltonians corresponding with electronic ground state problems contain $O(n^4)$ Pauli strings---whose row entries are zero except for a single easily retrievable entry \cite{Choo_2020}---it follows that they are efficiently computable. We now consider condition 2. The NQS ansatz, as defined in \cref{eqn:nqs_ansatz}, allows for easily retrievable state vector entries. Furthermore, the probabilities associated with $\abs{\psi_\theta}^2$ are determined by the individual entries of $\ket{\psi_\theta}$, so efficient access to these entries makes sampling, either approximate or exact, from this distribution possible.

The original NQS ansatz was the restricted Boltzmann machine (RBM), a graphical model designed to learn probability distributions \cite{carleo2017solving}. RBMs model unnormalized distributions and require an approximate Monte Carlo method like Metropolis-Hastings for efficient sampling. Moreover, these sampling methods involve significant computational overhead in order to generate uncorrelated samples, and therefore do not parallelize well \cite{zhao2021overcoming}. As a result, NQS development has generally moved away from RBMs toward autoregressive models that output conditional probabilities of spin configurations; these architectures generally follow the neural autoregressive quantum states structure from \cite{sharir2020deep}.

Consider an arbitrary fixed ordering of input spins $(x_1,\dots,x_n)\in\braces{0,1}^n$. We may construct an autoregressive ansatz using a set of real-valued functions: one phase function $\phi_\theta:\braces{0,1}^n\rightarrow \bracket{-\pi,\pi}$ and $n$ modulus functions $\log\psi_\theta^t:\braces{0,1}^j\rightarrow\bracket{0,1}$ outputting a sequence of autoregressive conditional probabilities \[\log p_\theta^j(x_1,\dots,x_j) = \log\operatorname{Pr}\paren{x_j=1\;|\;x_{k<j};\;\theta}.\]
From this family of functions the actual NQS ansatz may be constructed according to
\begin{equation}
    \log\ip{x|\psi_\theta} = \frac 1 2\sum_{j=1}^n p_\theta^j(x_{k\leq j})+ i\phi_\theta(x).
    \label{eqn:conditional_prob}
\end{equation}
Constructed from autoregressive conditional probabilities, this ansatz can only encode normalized quantum state vectors: furthermore, exact sampling from $\abs{\psi_\theta}^2$ may be done efficiently through qubit-by-qubit sampling of individual spins, using the conditional probabilities given by $p_\theta^1,\dots,p_\theta^n$.

The original autoregressive NQS ansatze used recurrent architectures \cite{hibat2020recurrent, sharir2020deep} to model the modulus functions $\{p_\theta^j\}_{j=1}^n$ with a single neural network, using an additional feedforward network for the phase function $\phi_\theta$. Although the first autoregressive ansatz applied specifically to \textit{ab initio} quantum chemistry used a family of separate networks for modulus functions \cite{barrett2022autoregressive}, the development of autoregressive NQS ansatze has shifted back toward the original vision of encoding all moduli within a single autoregressive network, for a more efficient representation of the ansatz.

\subsection{Variational Neural Annealing}
We give here a brief description of neural annealing, a training strategy that improves accuracy of NQS by regularizing the entropy of its ansatz distribution. A commonplace technique in reinforcement learning (RL) is to regularize the loss function by the entropy of the policy network---the probability distribution of actions that the RL agent will take. Entropy is an information-theoretic measurement of a distribution's randomness \cite{shannon1948information}, so such regularization encourages the model to explore a larger variety of actions during training. Otherwise, the model might default to a suboptimal policy with collapsed support inside the overall action space \cite{mnih2016asynchronous}. Since NQS parallels RL in such fundamental ways, \cite{hibat2021variational} first demonstrated how entropy regularization, referred to in this context as variational neural annealing (VNA), may be incorporated into the NQS paradigm. Utilizing the log-derivative trick here yields a stochastic estimator for the entropy gradient of $\ket{\psi_\theta}$:
\begin{align*}
    \nabla_\theta\paren{\underset{x\sim \abs{\psi_\theta}^2}{\mathbb{E}}\bracket{-\log\abs{\ip{x|\psi_\theta}}^2}}
    &= -2\operatorname{Re}\underset{x\sim \abs{\psi_\theta}^2}{\mathbb{E}}\bracket{\paren{1 + 2\operatorname{Re}\log \ip{x|\psi_\theta}}\nabla_\theta\log\ip{\psi_\theta|x}}.
    \numberthis
    \label{eqn:entropy_gradient}
\end{align*}

The derivation given in \cref{eqn:entropy_gradient} follows from elementary properties of the complex logarithm, complex conjugation, and from the fact that $\ket{\psi_\theta}$ is a complex-valued real function of $\theta$, as mentioned in \cref{sec:background_nqs}. If we choose to regularize \cref{eqn:cost_fn} by a small multiple $\beta$ of the ansatz entropy,
\begin{align*}
    \mathcal{L}(\theta) &= L(\theta) - \beta H_E(\ket{\psi_\theta})\\
    &= \underset{x\sim \abs{\psi_\theta}^2}{\mathbb{E}}\bracket{l_\theta(x) + \beta\log\abs{\ip{x|\psi_\theta}}^2},
    \numberthis
    \label{eqn:regularized_loss}
\end{align*}
then the gradient of \cref{eqn:regularized_loss} is given by
\begin{equation}
    \label{eqn:regularized_gradient}
    \nabla_\theta \mathcal{L}(\theta) = 2\operatorname{Re}\underset{x\sim \abs{\psi_\theta}^2}{\mathbb{E}}\bracket{\paren{l_\theta(x) + \beta(1 + 2\operatorname{Re}\log \ip{x|\psi_\theta}) - b}\nabla_\theta\log\ip{\psi_\theta|x}}.
\end{equation}
Analogously to the unregularized case, the normalization term $b$ is arbitrary, but typically taken to equal \cref{eqn:regularized_loss} \cite{hibat2021variational}. Utilizing a rate schedule that decays $\beta$ over the course of training, we may rely on this regularization to act akin to simulated annealing, preventing the NQS ansatz from becoming trapped in a sub-optimal local minimum without compromising accuracy of the minimal energy value obtained.

\subsection{Transformer As NQS Ansatz}
\label{subsection:transformers}

Transformer-based NQS ansatze \cite{shang2023solving} were introduced in part to improve upon an existing parallelizable autoregressive ansatz \cite{zhao2021overcoming} based on MADE \cite{germain2015made}, a feedforward network that preserves the autoregressive property by applying a set of fixed binary masks to its own weights, making it a strong choice for the modulus network. MADE outputs a full set of conditional probabilities for its given input sequence, so effective sampling requires $n$ full passes through the network, starting with a sequence of dummy spins and replacing them qubit-by-qubit with sampled configurations. The MADE ansatz demonstrates superior accuracy to CCSD techniques when applied to electronic ground state calculations \cite{zhao2023nnqsmade}.

MADE's simple structure and ease of parallelization does come at a price: the binary masks necessitate that a substantial portion of network weights do not contribute to the model's output, reducing MADE's expressiveness relative to its size. Moreover, as a feedforward network, the size of the MADE network is explicitly tied to the number of qubits in the NQS system and must grow with it. These hindrances motivate the search for more expressive and scalable autoregressive networks, like the transformer. First developed for natural language processing (NLP), the transformer \cite{vaswani2017attention} operates in direct contrast with recurrent models---which process input sequences entry-by-entry through multiple forward passes---by processing the entire sequence of inputs at once inside a single forward pass. The fundamental architectural component of the transformer is the multi-headed attention block, which processes the input sequences. The functional form of a single attention head is given as follows
\begin{equation}
    \operatorname{Attention}(X) = \operatorname{softmax}\paren{\frac{QK^T}{\sqrt{d_k}}}V, \text{ where } Q = XW_Q,\; K = XW_K,\; V=XW_V
    \label{eqn:self_attention}
\end{equation}
Here, a batch $X$ of (embedded) input vectors is projected by $W_Q$, $W_K$, and $W_V$ to batches $Q$, $K$, and $V$ of $d_k$-dimensional query, key, and value vectors. In practice, we choose $d_v=d_k$. After computing all possible inner products between queries and keys, a softmax function converts the inner products corresponding with a specific query into weights for the values: the result is referred to as a context vector. Several residual, layer norm, and feedforward layers perform additional processing on these context vectors following attention.

Informally, attention compares all pairs of entries in the input sequence in order to produce a single weighted context vector for each input. A multi-headed attention block independently operates several attention heads in tandem so the transformer can learn multiple forms of context simultaneously. The explicit term-by-term comparison of inputs not only parallelizes easily during training, but also allows attention to capture correlations between relatively distant inputs in the sequence, a feature distinguishing transformers from RNNs that rely on an iteratively updated internal memory vector to identify and preserve correlations between inputs. The initial transformer architecture is an encoder-decoder model, with only the decoder being truly autoregressive. In practice, the encoder and decoder are often used independently, and both can be made autoregressive with proper input masking.

The NQS-Transformer architecture \cite{wu2023nnqstransformer} applies the NQS ansatz QiankunNet \cite{shang2023solving}, which pairs a transformer modulus network with a feedforward phase network, to electronic ground state calculations at problem sizes beyond those tested with MADE. Because transformers explicitly learn correlations between inputs, and because the number of parameters given in a transformer is not intrinsically tied to the length of the input sequences---in this case, the number of qubits $n$ in the NQS problem---the transformer provides a suitable candidate for improved NQS scalability over MADE despite being a more sophisticated architecture. Nonetheless, the transformer still presents some room for improvement in scalability, particularly during inference. Regardless of how many parameters it possesses, the transformer will always require $O(n^2)$ time to perform a forward pass. This bottleneck occurs inside the attention heads, since attention explicitly and nonlinearly compares all pairs of inputs. This time complexity is unavoidable even for autoregressive attention, since input is so compared with every preceding input. The caching of keys and values during inference, known as KV caching \cite{pope2023efficiently}, increases efficiency by eliminating redundant calculations, and FlashAttention \cite{dao2022flashattention} and its successors have helped to significantly reduce the time needed to perform attention on GPU hardware. Nonetheless, quadratic time complexity is an inherent property of attention.

Since NQS requires multiple forward passes of its own ansatz at every iteration in order to create the data used to train the model, a model that possesses the transformer's advantages in training, but can perform inference in sub-quadratic time, would possess an inherent scalability advantage as an ansatz. This desire exists within the general field of LLMs beyond NQS, and several efforts have been made to reduce attention complexity through linear approximation of attention \cite{katharopoulos2020transformers}, dual-form RNNs that offer greater parallelization during training \cite{peng2023rwkv, sun2023retnet}, and state space models \cite{gu2023mamba}. These models show significant promise and it is natural to consider any one of them as a potential transformer alternative in NQS.

\subsection{Retentive Networks (RetNets)}
\label{subsection:retnet}
The Retentive Network (RetNet) constitutes one of several recurrent neural network architectures that have been developed as potential alternatives to transformers for use in LLMs \cite{sun2023retnet}. Sharing fundamental similarities with the RWKV model \cite{peng2023rwkv}, the RetNet distinguishes itself by its dual formulation as two separate, mathematically equivalent architectures for specific use in training and inference. Fundamentally, the RetNet is a recurrent model that can be collapsed into a mathematically equivalent parallel formulation. The parallel structure processes a sequence of inputs and calculates outputs, in quadratic time, within a single forward pass. In this form, RetNets train in a parallelizable manner analogous to transformers, without the concerns and inefficiencies that result from calculating the gradients of an RNN through multiple forward passes. At the same time, the recurrent formulation processes inputs iteratively, improving time complexity during inference. In direct contrast with KV-cached transformers, this iterative input processing executes in linear time, with constant additional memory constraints. A RetNet uses the same underlying weights in both its parallel and recurrent forms, and it is simple to switch between the two repeatedly within a single algorithm.

\begin{figure}
    \centering
    \begin{subfigure}{0.45\textwidth}
        \centering
        \includegraphics[scale=0.68]{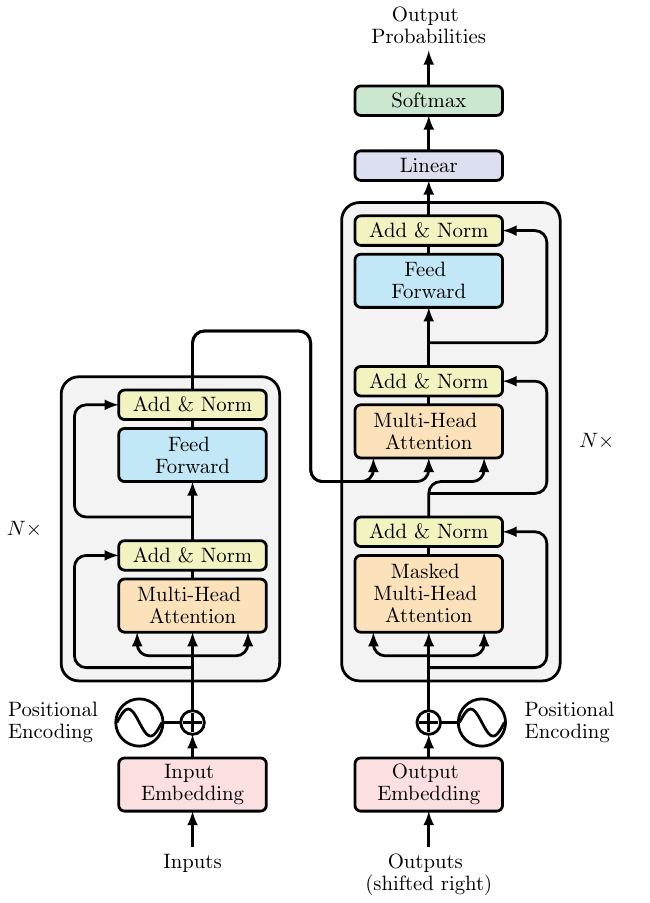}
        \caption{\em \small}
        \label{fig:transformer}
    \end{subfigure}
    \hfill
    \begin{subfigure}{0.45\textwidth}
        \centering
        \includegraphics[scale=0.28]{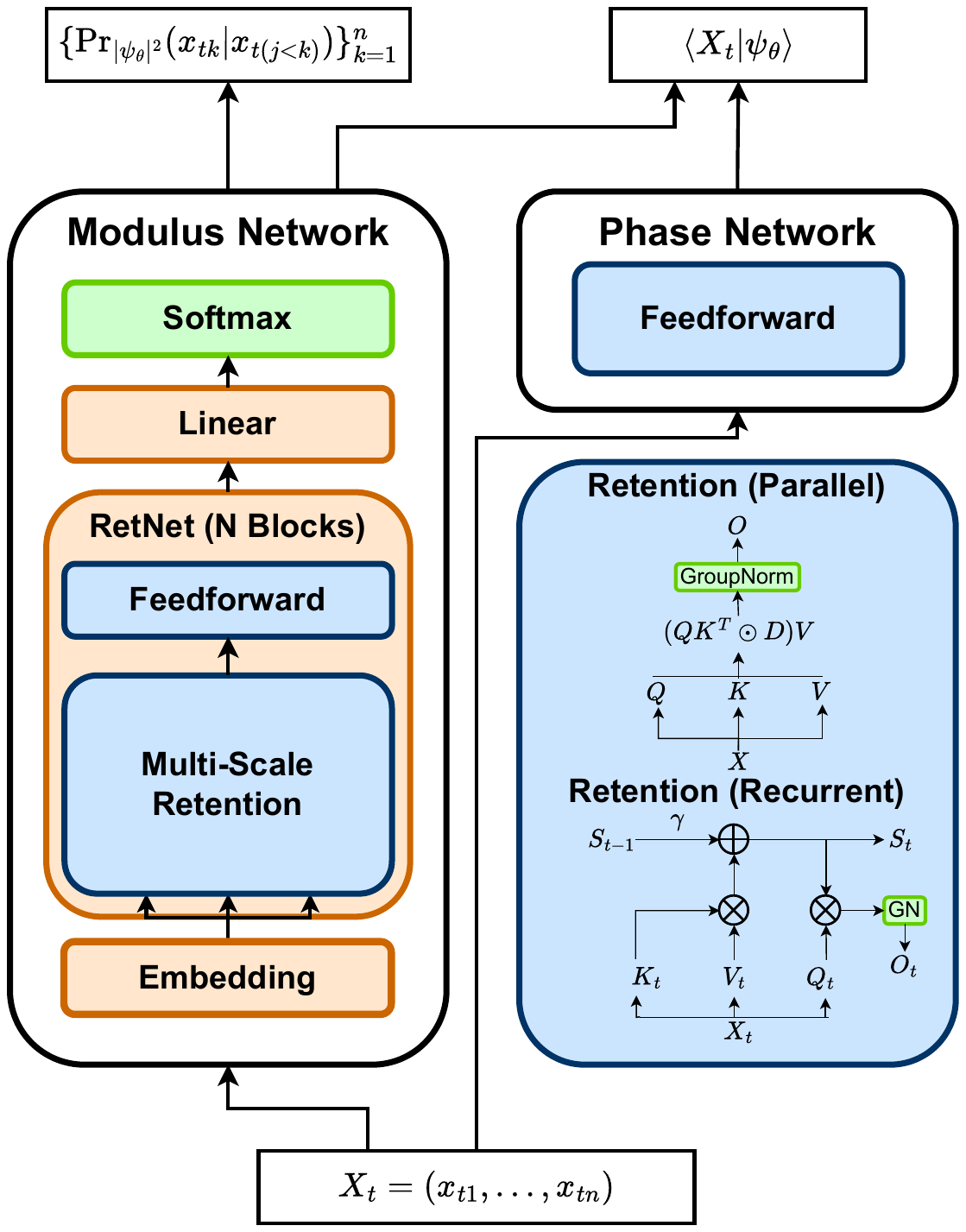}
        \caption{\em }
        \label{fig:retnet}
    \end{subfigure}
    \caption{(a)  A diagram of the transformer architecture originally featured in \cite{vaswani2017attention} reproduced with permission. (b) An NQS ansatz incorporating the RetNet architecture. We note the dual structure of the retention module, a proxy for attention that underpins the RetNet architecture \cite{sun2023retnet}. Retention may be computed in both parallel (left) and recurrent (right) forms: this dual formulation is what allows the RetNet to achieve the same ease of training as transformers, while performing inference as quickly as RNNs.}
\end{figure}

 As shown in \cref{fig:transformer}, each transformer block contains a multi-head attention layer followed by a feedforward layer, with residual connections and layer normalization following each block. A RetNet block, as depicted within \cref{fig:retnet}, is almost entirely identical to a decoder-only transformer block: the only significant architectural difference between the two is the RetNet's replacement of multi-head attention with multi-scale retention, a recurrent analogue of attention that makes the dual formulations of the RetNet possible \cite{sun2023retnet}. We also observe that RetNet performs two linear projections between retention and the feedforward layer, in contrast to transformer's single projection.

 A retention head operates similarly to an attention head: it uses learnable matrices $W_Q$, $W_K$, and $W_V$ to map a sequence $X$ of embedded input features to batches of queries, keys, and values. Additionally, a retention head possesses a vector $\theta\in\mathbb{R}^{\frac{d_m}2}$---where $d_m$ is the embedded input dimension---and a mask weight $\gamma\geq 0$, that define the following encoding matrix $\Theta$ and autoregressive mask matrix $D$:
 \begin{equation}
     \label{eqn:theta_and_D}
     \Theta_j = e^{ij\theta},\;\;\; D_{jk} = \begin{cases}
         \gamma^{j-k}, & j\geq k\\
         0, & j < k\\
     \end{cases}
 \end{equation}

These matrices serve together as a proxy for the repeated action of an invertible transformation matrix $A$ in a linear recurrent model. The constant $\gamma$ acts as the modulus for the complex eigenvalues of A---which is kept identical for all eigenvalues for ease of training---and the vector $\theta$ contains the eigenvalues' arguments. In practice, both $\gamma$ and $\theta$ are fixed parameters of the retention block.

Within retention, the matrix $\Theta$ adds a positional encoding to the queries and keys in a manner akin to xPos \cite{sun2022length}, while $D$ accounts for both the autoregressive masking and exponential decay of information across the input sequence \cite{sun2023retnet}. If we let $\overline{\Theta}$ represent the matrix whose entries are complex conjugates to those in $\Theta$, then we may express the parallel formulation of retention quite succinctly:
\begin{equation}
    \operatorname{Retention}(X) = \paren{QK^T\odot D}V, \text{ where } Q = (XW_Q)\odot\Theta,\; K = (XW_K)\odot\overline\Theta,\; V=XW_V
    \label{eqn:parallel_retention}
\end{equation}
RetNet follows the xPos convention of applying $\Theta$ to the query and key embeddings as if each of their rows contains the real and imaginary parts, as separate entries, for $\frac{d_m} 2$ complex numbers, but the succeeding operations operations treat $Q$ and $K$ as real-valued matrices.

Parallel retention depicts the clear analogy between RetNet and the transformer; fundamentally, however, RetNet is a recurrent architecture and may be expressed as such. Keeping $Q$, $K$, and $V$ defined as in \cref{eqn:parallel_retention}, we can recover retention from a recurrence on the rows $t=1,\dots,n$ of these matrices:
\begin{align*}
    & S_t = \gamma S_{t-1} + (K_t)^TV_t, \; S_0 = 0\\
    & \operatorname{Retention}(X_t) = Q_tS_t
    \numberthis
    \label{eqn:recurrent_retention}
\end{align*}

Within attention, the softmax component necessitates the explicit calculation of individual query-key products, resulting in $O(n^2)$ computational cost even with KV caching. In contrast, retention relates the queries, keys, and values in a purely linear way with respect to each query $Q_t$, allowing all the required information about previous keys and values to be stored as a single vector $S_{t-1}$. Qualitatively, this makes retention a form of linearized attention as described in \cite{katharopoulos2020transformers}. The retention-based analog for multi-head attention is gated multi-scale retention, which is generally analogous, but with a few caveats: several independent retention heads operate on the input sequence in tandem, each assigned its own separate, fixed scale parameter $\gamma$. The output of each head is independently normalized using GroupNorm, and then through the use of several learnable projection matrices---and a swish gate application for increased nonlinearity \cite{sun2023retnet}---the embedded input is integrated with the concatenated retention output and projected to a single output.

\section{Incorporating RetNet Ansatze and Neural Annealing into \textit{ab initio} NQS}
\label{sec:architecture}

Due to the architectural similarity between RetNets and transformers discussed in \cref{subsection:retnet}, one may consider simply replacing transformers with RetNets inside any larger deep learning pipeline. We propose that it is prudent to do so within the NQS-Transformer ansatz since RetNets have already demonstrated comparable performance to transformers on baseline NLP tasks \cite{sun2023retnet}. While more thorough research is likely necessary for determining the full extent to which the practical capabilities of these models match each other across all possible applications, when viewed as an analogy to NLP problems, NQS is much less sophisticated than typical use cases of these architectures. The vocabulary size is exceptionally small---hypothetically, only two spin choices per qubit, though in practice qubits are paired into orbitals with four distinct spin configurations. The sequence lengths, equalling half the number of qubits, are fixed for each training instance, so sequence lengths for even the largest systems considered would not translate to more than several sentences of text in NLP. Under these circumstances, RetNets appear to be a natural substitute for transformers.

Fig. \ref{fig:retnet} illustrates the architecture of our RetNet-based NQS ansatz, which is almost entirely analogous to both the MADE and transformer ansatze. The RetNet forms the backbone of the modulus network, while we still utilize a feedforward network for the phase component of the ansatz. In keeping with the transformer ansatz, the input sequence qubits are paired off to represent occupancies for entire orbitals, before being passed through a learnable embedding process into a higher-dimensional feature space. In order for the autoregressive property to be preserved, we append a universal start token to the beginning of every input sequence, and discard the final spin. Though the frequency matrix $\Theta$ already incorporates positional information within the RetNet, we still include a trainable positional encoding for expressiveness. After one or several RetNet blocks have processed the input sequence, a final linear layer is applied independently to each output of the RetNet sequence, at which point a softmax layer generates a four-dimensional probability vector corresponding with the two-qubit joint density for the corresponding orbital.

Following \cite{zhao2023nnqsmade}, we sample orbital occupancies in reverse order from their Jordan-Wigner ordering, to make use of the heuristic observation that the occupancies of higher-index orbitals are less likely to correlate with those of lower orbitals, making the reverse ordering better aligned with the nested conditional probabilities modeled by the autoregressive ansatz.

We also continue the practice from \cite{zhao2023nnqsmade} of constraining the modulus network to only sample from physically plausible state configurations. The number of electrons that can populate both spin-up and spin-down orbitals is fixed for each problem. Without explicit constraint, the ansatz will sample orbital occupancy configurations that correspond with different numbers of electrons, and are therefore physically infeasible. For each input sequence, a cumulative total is calculated across the occupancies of each orbital to guarantee the sequence's physical feasibility. If any sequences are found to be infeasible, masks are placed in the softmax layer of the corresponding model outputs to guarantee those sequences are not incorporated into the ansatz.

To maximize scalability, we utilize the recurrent formation of the RetNet for all forward passes for which backpropagation is not needed. Primarily, all Monte Carlo sampling is performed with the recurrent RetNet; we can successfully generate a full batch of samples with one full pass of the RetNet through an autoregressive generation process akin to a decoder model in NLP. Since it is possible in standard implementations to directly access, modify, and save the internal state vector of a recurrent RetNet, this generation can be accomplished with minimal forward passes of the network by copying and reusing internal state vectors of the model. This procedure of reusing internal state vectors is necessary since the ansatz generation process does not produce a full (redundant) set of samples. Instead, the process keeps a running tally of samples as they are generated orbital-by-orbital, only keeping one copy of each unique sample in order to save memory \cite{zhao2023nnqsmade}. Unlike the MADE and transformer ansatze, the structure of the RetNet ansatz serves it well for avoiding redundant calculations under this memory-efficient form of sampling.

Once the Monte Carlo samples are obtained, computing \cref{eqn:nqs_loss} and \cref{eqn:nqs_gradient} requires performing a substantial number of forward passes to generate the ansatz entries contributing to the numerator term $\ip{x|H|\psi_\theta}$ of each local energy sample. These forward passes correspond with bit flip operations applied to the actual Monte Carlo samples, as determined by the Pauli string terms in \cref{eqn:pauli_ham}. As discussed in \cref{sec:experiments}, it is possible to mitigate the number of forward passes needed by collecting all the unique bit flip operations among the terms of $H$, but these forward passes constitute the primary computational bottleneck of NQS in any case, and they do not contribute to the backpropagation. We rely on the recurrent form of the RetNet to generate these values. As observed in \cref{eqn:nqs_gradient}, the only ansatz entries that contribute to the backpropagation are those corresponding with the actual Monte Carlo samples, and we reserve the parallel RetNet for computing these entries. This configuration maximizes the scaling benefits of the RetNet relative to the transformer, for which we provide a basic analysis.

\subsection{Computational Cost Estimates for RetNets}
\label{subsection:flop_estimates}

Here, we give a simple computational cost estimate of the RetNet in the same vein as similar estimates made for transformers. By directly comparing these estimates, we identify a clear threshold ratio of problem-to-model-size past which the RetNet's inference time complexity outperforms the transformer. Since the actual time required to train a neural network depends on the specific hardware used, general algorithmic assessments of LLMs instead rely on floating point operation (FLOP) count estimates to measure compute requirements. The overall FLOP count for training a neural network is typically dominated by the number of operations needed to perform all forward and backward passes through the network during training, which scales naturally along input batch size and number of optimization steps. For every neural network, the FLOP count of a backward pass is approximately twice that of a forward pass, so it is natural to extrapolate the total FLOP count from that of a single forward pass operating on a single input instance \cite{kaplan2020scaling}. These counts are typically expressed as a function of the model's total number of trainable parameters, $N$, highlighting the direct effect of model size on compute requirements for a specific task.

\begin{table}[ht!]
    \centering
    \begin{tabular}{cccc}
        \toprule
        Layers&Parameter Count& Parallel RetNet FLOPs & Recurrent RetNet FLOPs\\
        \midrule
        QKV & $3n_\text{block}d_{\text{model}}d_\text{retn}$ & $6n_\text{block}d_\text{model}d_\text{retn}$ & $6n_\text{block}d_\text{model}d_\text{retn}$ \\
        Retention & 0 & $4n_\text{block}n_\text{seq}d_\text{retn}$ & $5n_\text{block}(d_\text{retn})^2$ \\
        Projection & $2n_\text{block}d_\text{model}d_\text{retn}$ & $4n_\text{block}d_\text{model}d_\text{retn}$ & $4n_\text{block}d_\text{model}d_\text{retn}$ \\
        Feedforward & $2n_\text{block}d_\text{model}d_\text{ff}$ & $4n_\text{block}d_\text{model}d_\text{ff}$ & $4n_\text{block}d_\text{model}d_\text{ff}$ \\
        \midrule
        Totals: & $N=2n_\text{block}d_\text{model}(2.5d_\text{retn} + d_\text{ff})$ & $2N + 4n_\text{block}n_\text{seq}d_\text{retn}$ & $2N + 5n_\text{block}(d_\text{retn})^2$ \\
        \bottomrule \\
    \end{tabular} 
    \caption{ Parameter and FLOP count estimates for an $n_\text{block}$-block RetNet with a single retention head, acting on a single token within an input sequence of length $n_\text{seq}$. Following convention, we express each FLOP count in terms of the number $N$ of trainable parameters in the model, and we ignore the contributions of nonlinearities, biases, normalizations and other sub-leading terms when making these estimates.}
    \label{tab:FLOP_count}
\end{table}

In \cref{tab:FLOP_count} we give FLOP count estimates for both the parallel and recurrent RetNet formulations, specifically what is needed to compute one forward pass for a single embedded input token through $n_\text{block}$ identically-sized RetNet blocks. We may directly compare these cost estimates with similar estimates made for the transformer network in \cite{kaplan2020scaling}. Following the conventions of \cite{kaplan2020scaling}, we consider only the leading terms of each count, which generally come from projection layers, the retention mechanism, and the feedforward layers, while discarding the contribution of sub-leading costs like nonlinearities, biases, normalizations, and the positional encoding and autoregressive masking given by $\Theta$ and $D$. The total count $N$ of trainable RetNet parameters is then approximately determined by the number $n_\text{block}$ of RetNet blocks and the sizes of the projection matrices within each block: the three projection matrices $W_Q$, $W_K$, and $W_V$, two projection matrices operating on the output produced by the retention mechanism, and the one-layer feedforward component. These counts are determined by the dimension of the model's embedded input, $d_\text{model}$; the dimension of the retention component $d_\text{retn}$, which is typically taken to equal $d_\text{model}$; and the dimension $d_\text{ff}$ of the feedforward component, which typically equals $4d_\text{model}$. The actual retention mechanism, which is the only place where the parallel and recurrent RetNets differ, does not contribute significantly to the model's parameter count, but does contribute to the forward pass FLOP estimate.

Projections layers, like feedforward layers, require approximately twice as many FLOPs as layer weights to perform a forward pass. As further broken down in \cref{tab:FLOP_count}, it therefore follows that the RetNet, like the transformer \cite{kaplan2020scaling}, performs a forward pass in approximately $2N$ FLOPs, plus the number of FLOPs required to perform retention. For parallel retention, this additional term equals $4n_\text{block}n_\text{seq}d_\text{retn}$, while in the case of recurrent retention, this term is given by $5n_\text{block}(d_\text{retn})^2$. These counts are given for a single retention head: in the case of multi-scale retention, the count remains relatively unchanged, since each head is projected to a fraction of $d_\text{retn}$. We also note, as shown in \cref{fig:retnet}, that since neural networks typically only use a small number of feedforward layers to process output from the RetNet/transformer blocks, the FLOP contribution from this post-processing can be absorbed into this $2N$ term determined by the number of network weights.

The FLOP count for the parallel RetNet is therefore entirely analogous to that of the transformer \cite{kaplan2020scaling}, with a small caveat: a RetNet block performs two post-retention projections, while a transformer only performs one projection post-attention, so a parallel RetNet will require $2n_\text{block}d_\text{model}d_\text{retn}$ more FLOPs per token to perform a forward pass than a transformer of equivalent dimension. This difference is generally quite small relative to the total FLOP count, especially when in the context of NQS, where we deliberately keep use of the parallel RetNet to a minimum. In any case, the per-token time complexity of both models depends on the input sequence length $n_\text{seq}$ due to the query-key products in \cref{eqn:self_attention} and \cref{eqn:parallel_retention}. For NQS, this length typically equals half the number of qubits; nonetheless, this component ensures that the forward pass time complexity of both models, applied to an entire input sequence, exhibits quadratic scaling with respect to sequence length.

When comparing the time complexity of the recurrent RetNet to the transformer, we do take into account the slight difference in size between the two models. At the same time, the per-token forward pass FLOP count of the recurrent RetNet does not depend on $n_\text{sec}$, meaning there exists a threshold past which the parallel RetNet operates using fewer FLOPs than the transformer, as identified by the condition \begin{equation}
4n_\text{block}n_\text{seq}d_\text{retn} >5n_\text{block}(d_\text{retn})^2 + 2n_\text{block}d_\text{model}d_\text{retn}.
    \label{eqn:scaling_threshold_retnet}
\end{equation}

Under the standard assumption that $d_\text{retn}=d_\text{model}$, this condition is satisfied if and only if
\begin{equation}
    \boxed{n_\text{seq} > 1.75d_\text{model}}
\end{equation}
In the case of LLMs, for which the size of $d_\text{model}$ typically falls within a range of several thousand \cite{zhao2023survey}, these findings significantly restrict the sequence lengths for which RetNets are expected to outperform transformers at inference. As explored in \cite{wu2023nnqstransformer} and in \cref{sec:experiments}, NQS can operate effectively using much smaller model sizes, yielding a much wider array of problems for which \cref{eqn:scaling_threshold_retnet} is satisfied. We note that typical scaling analysis of transformers \cite{kaplan2020scaling} only approximates the per-token forward pass FLOP count by $2N$, under the assumption that the additional attention cost is negligible if $12d_\text{model}\gg n_\text{seq}$, which is the case for most LLMs. This argument suggests a further restriction on the practical utility of RetNets as a general replacement for transformers, but in the case of NQS, the substantial number of forward passes performed at each optimization step---alongside the smaller model sizes---ensures that any time complexity advantage held by the RetNet will accumulate to a larger fraction of the overall compute cost than a typical LLM training framework.

\subsection{Utilizing VNA in Electronic Ground State NQS}

While aiming to improve the time complexity of autoregressive NQS at the architectural level, we also explore incorporating variational neural annealing to enhance NQS accuracy relative to model size. Variational Neural Annealing was first applied to NQS in direct analogy to standard simulated annealing techniques, to aid in solving combinatorial optimization problems modeled as Ising Hamiltonians \cite{hibat2021variational}. These demonstrations were made using the RNN ansatz from \cite{hibat2020recurrent}, but VNA may be applied in the same manner to other autoregressive ansatz, in particular to the MADE, transformer, and RetNet ansatze. We also propose electronic ground state problems as a natural use case for VNA. The true electronic ground state for a single molecule typically resides in a small subspace of the overall ansatz state space: during training, the model state distribution can often collapse into too small of a subspace, from which the model struggles to learn a viable path toward the true ground state. VNA encourages the ansatz to search for the ground state from within a larger fraction of the state space, preventing this mode collapse from occurring.

We recall from \cref{eqn:regularized_loss} that under VNA, the perturbation of the NQS loss by the ansatz entropy is controlled through the regularization parameter $\beta$, which must be decayed to zero according to some choice of rate schedule during the course of training. The original VNA paradigm \cite{hibat2021variational}, while ultimately agnostic to the choice of schedule made here, primarily decayed $\beta$ linearly in a direct analog to simulated and quantum annealing. We propose using a more general polynomial annealing schedule to vanish $\beta$ instead: \begin{equation}
\beta(t) = \paren{1 - \frac t T}^r,
    \label{eqn:vna_annealing}
\end{equation} where $T$ equals the total number of annealing iterations and $r>1$ controls the desired decay rate. We utilize this annealing schedule as it is superlinear for the initial part of training, where precision is less important since ansatz entropy is still quite high; this serves to decrease computational time in the first half of training. At the same time, the schedule is sublinear for the majority of training time, allowing the final portion of the annealing to proceed slowly to ensure greater accuracy. Empirical testing suggests that $r$ should not be made too large, since decreasing $\beta$ too quickly can result in poor accuracy. We observe in practice that setting $r=4$ results in a suitable annealing behavior for most problems.

We find in practice that it is most effective to begin annealing after small number of warmup steps, which we generally take to be $0.04T$ for our experiments in \cref{sec:experiments}. In these experiments, we illustrate how variational neural annealing significantly improves the robustness---and, indirectly, the practically attainable accuracy---of both the transformer and RetNet ansatze. More specifically, \cref{tab:vna_comparison} illustrates how for a specific model size, learning rate schedule, and sample schedule, the introduction of VNA alone can improve the training accuracy for multiple NQS ansatze to baseline standards. These results seem to indicate that state of the art levels of accuracy may be attainable using smaller model sizes than those previously applied to electronic structure problems.

\section{Experimental Results}
\label{sec:experiments}

We describe our experimental framework for demonstrating the capability of the RetNet ansatz from \cref{sec:architecture}. We also demonstrate how adding neural annealing can increase the ease with which the MADE, transformer, and RetNet ansatze can identify ground state energies. For all our testing, we follow common practice for storing and operating on the Pauli string terms in our Hamiltonian \cite{wu2023nnqstransformer, zhao2023nnqsmade}. More specifically, each Pauli string is split into two strings: one containing bit flip indices, defined by the placement of Pauli $X$ and $Y$ matrices, and one containing sign flip indices, which are defined by placement of the $Y$ and $Z$ matrices. This decomposition allows for efficient batch calculation of local energies---we observe that each local energy calculation requires obtaining $\ip{x|H|\psi_\theta}$ for some sample $x\sim\abs{\psi_\theta}^2$, which is the product of row $x$ of $H$ with the ansatz $\ket{\psi_\theta}$. The sign flip indices are used to retrieve the nonzero entries in row $x$ for each term in \cref{eqn:pauli_ham}. At the same time, the bit flip indices calculate the corresponding entries of $\ket{\psi_\theta}$ that multiply with those of $H$ to create $\ip{x|H|\psi_\theta}$.

Additionally, we follow the data-parallelized training setup from \cite{zhao2023nnqsmade}, which has since become standard practice. We parallelize the local energy calculation across several GPUs, with each GPU receiving its own copy of both the ansatz network and the Hamiltonian. Local energy values are collected across GPUs in order to estimate loss values, and gradients are similarly collected for optimization as well. We follow the practice from \cite{wu2023nnqstransformer} of storing only the unique bit flip index strings present among all Hamiltonian terms, for increased Hamiltonian storage efficiency on each GPU. Likewise, we also follow the practice of slowly increasing the number of samples used to estimate \cref{eqn:nqs_gradient} over the course of training---starting with $10^4$ samples per iteration and reaching $10^{12}$ samples for the final 10\% of training---as a measure of balancing accuracy with speed. At the end of training, we return the lowest loss value obtained from the following process: t each update of the loss function, we compare the new loss value with the lowest value previously obtained: using the sample variance for the new estimate, we accept the new loss as the lowest value if the previous lowest value exceeds its 95\% upper confidence bound. This practice is intended to reduce the outside chance that the algorithm conflates favorable statistical noise with genuine improvement in accuracy, especially when fewer samples are used at the start of training. With the number of samples used by the end, and the fact that the variance of \cref{eqn:nqs_loss} tends to 0 as it approaches the ground state \cite{zhao2021overcoming}, it is virtually guaranteed in practice that the minimal value returned is accurate to at least seven or eight decimal places.

\begin{table}[ht!]
    \centering
    \small
    \begin{tabular}{ccccccccc}
        \toprule
        Molecule & $n_{\text{qubits}}$ & $N_E$ & Paulis & CCSD & MADE & Transformer & RetNet & FCI \\
        \midrule
        H\textsubscript{2}O & 14 & 10 & 1390 & $-75.0154$ & $-\textbf{75.0155}$ & $-\textbf{75.0155}$ & $-\textbf{75.0155}$ & $-75.0155$ \\
        N\textsubscript{2} & 20 & 14 & 2591 & $-107.6560$ & $-107.6598$ & $-\textbf{107.6599}$ & $-\textbf{107.6599}$ & $-107.6602$ \\
        O\textsubscript{2} & 20 & 16 & 2879 & $-147.7477$ & $-\textbf{147.7502}$ & $-\textbf{147.7502}$ & $-\textbf{147.7502}$ & $-147.7502$ \\
        H\textsubscript{2}S & 22 & 18 & 9558 & $-394.3546$ & $-394.3543$ & $-\textbf{394.3545}$ & $-394.3544$ & $-394.3546$ \\
        PH\textsubscript{3} & 24 & 18 & 24369 & $-338.6982$ & $-338.6970$ & $-\textbf{338.6971}$ & $-338.6968$ & $-338.6984$ \\
        LiCl & 28 & 20 & 24255 & $-460.8476$ & $-460.8490$ & $-460.8492$ & $-\textbf{460.8493}$ & $-460.8496$ \\
        Li\textsubscript{2}O & 30 & 14 & 20558 & $-87.8855$ & $-87.8668$ & $-87.8817$ & $-\textbf{87.8848}$ & $-87.8927$ \\
        \bottomrule \\
    \end{tabular}
    \caption{ Ground state energies (in Hartree) of a selection of small molecules, identified by VNA-augmented MADE, transformer, and RetNet ansatze. CCSD \cite{zhao2023nnqsmade} and FCI \cite{wu2023nnqstransformer} baselines shown for comparison.}
    \label{tab:retnet_comparison}
\end{table}

We find in practice that the greatest memory bottleneck during training comes from the multiple forward passes of the ansatz required to calculate all flipped sample states. Each sampled state $x\sim\abs{\psi_\theta}^2$ necessitates one forward pass to calculate $\ip{x|\psi_\theta}$, in addition to one forward pass for every single unique bit flip variation contained in $H$, as applied to $x$. The memory requirements to process all these forward passes in parallel grow with problem size at a much faster pace than they do for any other aspect of the architecture. To mitigate this, we split forward passes of flipped sample states into batches, and in doing so we find that fixing even a small number of batches, even as low as 8, can lead to a substantial decrease in required GPU memory without a significant qualitative increase in execution time. Since these ansatz values are all being obtained in service of calculating local energies, which are all treated as constant values when calculating \cref{eqn:nqs_gradient}, no additional gradient accumulation is needed to perform this batching.

To help decrease execution time, we recall from previous discussions of \cref{eqn:nqs_gradient} that calculation of the actual loss function estimate is not necessary to estimate gradients; though \cref{eqn:nqs_loss} is typically incorporated into \cref{eqn:nqs_gradient} as the normalizer term $b$ to help with variance reduction, it is not at all necessary that $b$ equal the loss value. Since a reasonable estimate for $\cref{eqn:nqs_loss}$ is likely to help reduce variance reasonably well, we do not directly calculate \cref{eqn:nqs_loss}---or \cref{eqn:regularized_loss} in the case of neural annealing---for the majority of training iterations. Instead, for the first 90\% of training, we only calculate loss values at regular intervals, utilizing the same loss value as a gradient normalizer for all intervening optimization steps. We find in practice that reusing loss function values can decreases execution time without sacrificing accuracy. To reduce training time, we also discard samples that occur exactly once at each stage of autoregressive sampling, since they are unlikely to make a significant contribution to the overall loss in cases where the number of non-unique samples is a sufficiently large fraction of the search space. It is worth emphasizing that this practice most likely needs to be discarded at scale, where the number of initial samples taken is too small a fraction of the search space to guarantee a significant number of unique samples to occur more than once, especially early in training when $\abs{\psi_\theta}^2$ is expected to be quite uniform.

\begin{table}[ht!]
    \small 
    \centering
    \begin{tabular}{ccccc c}
        \toprule
        Molecule & Transformer & Transformer (VNA) & RetNet & RetNet (VNA) & FCI \\
        \midrule
        H\textsubscript{2}O & $\textbf{-75.0155}$ & $\textbf{-75.0155}$ & $-74.9951$ & $\textbf{-75.0155}$ & $-75.0155$ \\
        N\textsubscript{2} & $-107.6245$ & $\textbf{-107.6599}$ & $-107.5922$ & $\textbf{-107.6599}$ & $-107.6602$ \\
        O\textsubscript{2} & $-147.6833$ & $\textbf{-147.7502}$ & $-147.6974$ & $\textbf{-147.7502}$ & $-147.7502$ \\
        H\textsubscript{2}S & $-394.3544$ & $\textbf{-394.3545}$ & $-394.3531$ & $-394.3544$ & $-394.3546$ \\
        PH\textsubscript{3} & $-338.6965$ & $\textbf{-338.6971}$ & $-338.6826$ & $-338.6968$ & $-338.6984$ \\
        LiCl & $-460.8418$ & $-460.8492$ & $-460.8314$ & $\textbf{-460.8493}$ & $-460.8496$ \\
        Li\textsubscript{2}O & $-87.8521$ & $-87.8817$ & $-87.8254$ & $\textbf{-87.8848}$ & $-87.8927$ \\
        \bottomrule \\
    \end{tabular}
    \caption{The effect of quartic-schedule variational neural annealing on both the transformer and RetNet ansatze. The two ansatze share the exact same architectural layout---number of hidden layers, width of hidden layers, input embedding dimension---and are both trained using Adam for 25000 iterations under a cosine annealing learning rate schedule. FCI ground truth values included for reference.}
    \label{tab:vna_comparison}
\end{table}

In \cref{tab:retnet_comparison}, we give a demonstration---on a collection of small molecules---of VNA-augmented MADE, transformer, and RetNet ansatze, in comparison with CCSD and full configuration interaction (FCI) ground truth values. Depending on ansatz type and problem size, these examples were generated using groups of four NVIDIA L4 or T4 GPUs. The LLM ansatze both contain a single decoder block with a hidden dimension of 16 and feedforward layer dimension of 64, while the MADE ansatz consists of two 64-dimensional hidden layers. All ansatze have a phase network with two 64-dimensional hidden layers. All models were trained with Adam for 25,000 steps, with a starting initial learning rate of $2.5\times 10^{-3}$, achieved after a warmup period equaling 4\% of the total number of steps, that was then annealed along a cosine schedule \cite{loshchilov2016sgdr} down to $5\times 10^{-8}$. We also perform neural annealing on a quartic schedule starting at 5\% of the total number of training steps. Under this setup, we find that all models perform comparably well, with RetNet and transformer essentially performing in lockstep. One important factor to highlight is that the transformer ansatz has been demonstrated in \cite{wu2023nnqstransformer} to achieve greater accuracy on these problems than what is listed here, but those results were obtained using significantly larger models that were trained for a greater number of iterations. For ease of making a direct comparison between models of similar dimension, and to help illustrate the beneficial effects of VNA on NQS accuracy with respect to model size, we choose to highlight results obtained with a much smaller model. For further ease of training, the number of unique samples used to generate local energy estimates was capped at 8000 at every stage of training for all molecules; in practice, such a cap is a user-specified compromise between training time and final accuracy.

The benefits brought to NQS by neural annealing may be more easily seen through an ablation test, as in \cref{tab:vna_comparison}. Here, we demonstrate the capability of VNA to improve training accuracy for both the transformer and RetNet ansatze. In all cases, our models are trained with Adam for 25,000 steps, with an initial learning rate of $2.5\times 10^{-3}$ that is cosine-annealed down to $5\times 10^{-8}$, with an initial warm-up equalling 4\% of total steps---an identical configuration to the tests used to populate \cref{tab:retnet_comparison}. We present the results of both models with and without a quartic-schedule neural annealing, initialized at 5\% of the total step number. For both models, the results are quite clear: without annealing, this training setup would be entirely inadequate, but annealing allows both models to reach a greater order of magnitude in accuracy in the same amount of time. These results appear to indicate that the incorporation of variational neural annealing might constitute a crucial step toward making NQS a practical black box electronic ground state solver.

\begin{table}[ht!]
    \small 
    \centering
    \begin{tabular}{ccccc}
        \toprule
        Molecule  & MADE (No VNA) & MADE (VNA) & MADE (Baseline) \cite{zhao2023nnqsmade} & FCI \\
        \midrule
        H\textsubscript{2}O & $-74.9992$ & $\textbf{-75.0155}$ & $\textbf{-75.0155}$ & $-75.0155$ \\
        N\textsubscript{2} & $-107.6338$ & $\textbf{-107.6598}$ & $-107.6568$ & $-107.6602$ \\
        O\textsubscript{2} & $-147.7352$ & $\textbf{-147.7502}$ & $-147.7500$ & $-147.7502$ \\
        H\textsubscript{2}S & $-394.3540$ & $-394.3544$ & $\textbf{-394.3546}$ & $-394.3546$ \\
        PH\textsubscript{3} & $-338.6896$ & $-338.6978$ & $\textbf{-338.6982}$ & $-338.6984$ \\
        LiCl & $-460.8458$ & $\textbf{-460.8490}$ & $-460.8481$ & $-460.8496$ \\
        Li\textsubscript{2}O & $-87.8437$ & $\textbf{-87.8859}$ & $-87.8856$ & $-87.8927$ \\
        \bottomrule \\
    \end{tabular}
    \caption{Depiction of how neural annealing improves accuracy of the MADE ansatz, even producing results exceeding baseline \cite{zhao2023nnqsmade} values. FCI ground truth values included for reference.}
    \label{tab:made_vna_comparison}
\end{table}

Not only does VNA improve the ability of smaller models to obtain more accurate results in less time, it also serves to help bridge the gap in known accuracy between the MADE ansatz and the more recent LLM-derived models. As demonstrated in \cref{tab:retnet_comparison}, while MADE ansatze do not necessarily outperform transformer and RetNet models of equivalent dimension under VNA, they nonetheless perform to comparable levels of accuracy. Furthermore, these new levels of accuracy can often exceed known baseline capabilities \cite{zhao2023nnqsmade} of the MADE architecture on electronic ground state calculations. Table \ref{tab:made_vna_comparison} depicts the VNA-augmented MADE ansatz surpassing these baselines in several significant instances, while also showing how the same MADE architecture without neural annealing would not be able to reach accuracy levels close to these under the same training setup. To better highlight the effect of neural annealing on the overall accuracy of the MADE ansatz, we obtained these results from a variety of hyperparameter configurations similar to those used for \cref{tab:retnet_comparison}, with the larger molecule problems trained for up to 50,000 iterations---the lithium oxide (Li\textsubscript{2}O) example was generated using a hidden layer dimension of 128 and a maximum unique sample cutoff of 16,000 to better improve model accuracy at this size.

\section{Conclusion}
\label{sec:conclusions}

In this paper, we have introduced a novel autoregressive NQS ansatz, based on retentive networks, as a potential alternative to existing state-of-the-art transformers. As a result of their dual parallel and recurrent formulations, RetNets exhibit many of the fundamental benefits of transformers, while at the same time performing inference with far better scalability than transformers. Since training an NQS ansatz requires generating significant data using the ansatz's own output, this improvement in scaling represents an additional step toward improving the practicality of NQS as a black box solver. Through a basic FLOP count comparison between RetNets and transformers, we have identified a suitable threshold ratio of problem size to model dimension past which RetNets will perform inference with improved time complexity compared to transformers of equivalent dimension. We discussed the application of variational neural annealing to electronic ground state problems, and introduced a simple annealing schedule for effective training.

Using this neural annealing schedule, we demonstrated that the RetNet reaches levels of accuracy comparable to state of the art ansatze for a slate of molecules corresponding with small and moderately sized NQS problems. We do note that the scale of these experiments is, to an extent, limited in comparison with \cite{wu2023nnqstransformer}, due to our comparatively smaller level of access to highly parallelizable multi-GPU hardware. Nonetheless, these results indicate that RetNet ansatze exhibit comparable levels of performance to transformer ansatze of similar sizes, when trained under similar conditions. Furthermore, we demonstrated how neural annealing improves the general trainability of autoregressive NQS ansatze on these electronic structure problems, by showing how the application of neural annealing alone improves the accuracy attainable by MADE, transformer, and RetNet ansatze. In the case of transformers and RetNets, our choice of model sizes in these experiments indicates that autoregressive NQS may be effectively performed using smaller models than previously believed, bringing the ratio of problem-to-model size needed for effective training closer to the point past which RetNets perform more efficiently as NQS ansatze. In the case of MADE, neural annealing was shown to improve its capabilities to a level of accuracy beyond what was previously known. Taken together, these results show a lot of promise for employing neural annealing as a general training strategy within a scalable NQS-based black box solver for electronic structure problems.

With regard to future work, one important observation to make is that NQS is a fundamentally problem-agnostic architecture, and it would be most helpful to gain a greater understanding of how both the transformer and Retnet ansatze perform when applied to NQS problems beyond electronic structure calculations. Likewise, NQS is also a model-agnostic architecture, and it is our hope that this work inspires further exploration of novel autoregressive NQS ansatze. Additionally, a more comprehensive study of neural annealing schedules would be immensely helpful for better understanding the amount of improvement that VNA can provide in practice. We also believe these results may renew interest in exploring MADE, a far simpler architecture than transformers or RetNets, as a competitive NQS ansatz for electronic structure problems.

\section*{Acknowledgements}
The authors would like to thank Lukas Scheiwiller and Tianchen Zhao for helpful discussions. JS and SV acknowledge support from the Michigan Institute for Data and AI in Society (MIDAS) and the Michigan Translational Research and Commercialization (MTRAC) Innovation Hub for Advanced Computing Technologies.

\bibliographystyle{plain}
\bibliography{references}
\end{document}